\documentclass[final]{cvpr}

\usepackage{times}
\usepackage{epsfig}
\usepackage{graphicx}
\usepackage{amsmath}
\usepackage{amssymb}
\usepackage{lipsum}
\usepackage{bm}
\usepackage{booktabs}
\usepackage{multirow}
\newcommand{\tabincell}[2]{\begin{tabular}{@{}#1@{}}#2\end{tabular}}
\usepackage{arydshln}
\usepackage{verbatim}
\usepackage{textpos}  
\usepackage{tablefootnote}
\usepackage[flushleft]{threeparttable}

\usepackage[pagebackref=true,breaklinks=true,colorlinks,bookmarks=false]{hyperref}

\def\cvprPaperID{686} 
\def\confYear{CVPR 2021}

\begin{document}

\title{CoLA: Weakly-Supervised Temporal Action Localization with \\Snippet Contrastive Learning}

\author{Can Zhang$^{1}$, Meng Cao$^{1}$, Dongming Yang$^{1}$, Jie Chen$^{1,2}$, Yuexian Zou$^{*1,2}$\\
$^{1}$School of Electronic and Computer Engineering, Peking University $^{2}$Peng Cheng Laboratory\\
{\tt\small \{zhangcan, mengcao, yangdongming, zouyx\}@pku.edu.cn; chenj@pcl.ac.cn}
}

\maketitle
\pagestyle{empty}
\thispagestyle{empty}

\begin{textblock*}{.8\textwidth}[.5,0](0.5\textwidth, -.20\textwidth)
\centering
{\textbf{Code: \url{https://github.com/zhang-can/CoLA}}}
\end{textblock*}

\begin{abstract}
Weakly-supervised temporal action localization (WS-TAL) aims to localize actions in untrimmed videos with only video-level labels. Most existing models follow the ``localization by classification" procedure: locate temporal regions contributing most to the video-level classification. Generally, they process each snippet (or frame) individually and thus overlook the fruitful temporal context relation. Here arises the single snippet cheating issue: ``hard" snippets are too vague to be classified. In this paper, we argue that learning by comparing helps identify these hard snippets and we propose to utilize snippet \textbf{Co}ntrastive learning to \textbf{L}ocalize \textbf{A}ctions, \textbf{CoLA} for short. Specifically, we propose a Snippet Contrast (SniCo) Loss to refine the hard snippet representation in feature space, which guides the network to perceive precise temporal boundaries and avoid the temporal interval interruption. Besides, since it is infeasible to access frame-level annotations, we introduce a Hard Snippet Mining algorithm to locate the potential hard snippets. Substantial analyses verify that this mining strategy efficaciously captures the hard snippets and SniCo Loss leads to more informative feature representation. Extensive experiments show that CoLA achieves state-of-the-art results on THUMOS'14 and ActivityNet v1.2 datasets. 
\end{abstract}

\vspace{-6pt}
\section{Introduction} \label{sec:intro}

Temporal action localization (TAL) aims at finding and classifying action intervals in untrimmed videos. It has been extensively studied in both industry and academia, due to its wide applications in surveillance analysis, video summarization and retrieval~\cite{vishwakarma2013survey,lee2012discovering,ma2005generic}, \emph{etc}. Traditionally, fully-supervised TAL is labor-demanding in its manual labeling procedure, thus weakly-supervised TAL (WS-TAL) which only needs video-level labels has gain popularity. 

\begin{figure}[t]
\begin{center}
\includegraphics[width=1\linewidth]{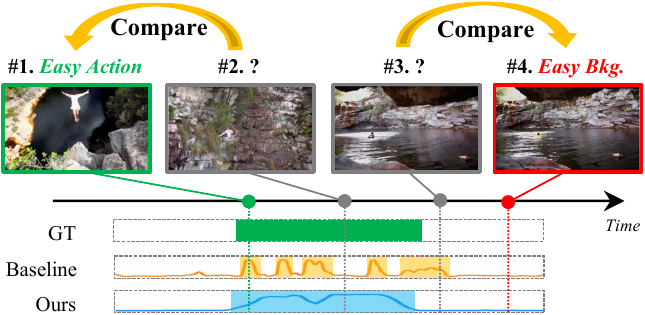}
\end{center}
 \caption{Which category do the two selected snippets (\#2, \#3) belong to? It is difficult to tell when evaluating independently and they are actually misclassified in \textit{baseline} (We plot the one-dimensional T-CAS for \textit{CliffDiving} and the thresholded results). By contrast, learning by comparing helps identify them: \#2 snippet (person falling down) is inferred to be the action snippet by making a comparison with \#1 ``easy action" (different camera views of the \textit{CliffDiving} action); The inference of \#3 snippet is also rectified after the comparison with \#4 ``easy background" snippet.}
\label{fig:intro}
\end{figure}

Most existing WS-TAL methods~\cite{wang2017untrimmednets,nguyen2018weakly,paul2018w,narayan20193c,lee2020background} employ the common attention mechanism or multiple instance learning formulation. Specifically, each input video is divided into multiple fixed-size non-overlapping snippets and the snippet-wise classifications are performed over time to generate the Temporal Class Activation Map/Sequence (T-CAM/T-CAS)\cite{nguyen2018weakly,shou2018autoloc}. The final localization results are generated by thresholding and merging the class activations. For illustration, we consider the na\"ive case where the whole process is optimized with a single video-level classification loss and we treat this pipeline as \textit{baseline} in our paper.

In absence of frame-wise labels, WS-TAL suffers from the \textit{single snippet cheating issue}: indistinguishable snippets are easily misclassified and hurt the localization performance. To illustrate it, we take \textit{CliffDiving} in Figure~\ref{fig:intro} as an example. When evaluated individually, two selected snippets (\#2, \#3) seem ambiguous and are misclassified: 1) the \#2 snippet is incorrectly categorized, thus breaking the time intervals; 2) the \#3 snippet is misidentified as an action in \textit{baseline}, resulting in inaccurately extended action interval boundaries. \textit{How to address the single snippet cheating issue?} Let's revisit the case in Figure~\ref{fig:intro}. By comparing snippets of interest with those ``easy snippets" which can be classified effortlessly, action and background can be distinguished more easily. For example, the \#2 snippet and the \#1 easy action snippet are two different views of a man falling-down process in ``CliffDiving". The \#3 snippet is similar to the \#4 easy background snippet and can be easily classified as the background class. In light of this, we contend that \textit{localizing actions by contextually comparing offers a powerful inductive bias that helps distinguish hard snippets}. Based on the above analysis, we propose an alternative, rather intuitive way to address the single snippet cheating issue -- by conducting \textbf{Co}ntrastive learning on hard snippets to \textbf{L}ocalize \textbf{A}ctions, \textbf{CoLA} for short. To this end, we introduce a new Snippet Contrast (SniCo) Loss to refine the feature representations of hard snippets under the guidance of those more discriminative easy snippets. Here these ``cheating" snippets are named hard snippets due to their ambiguity. 

This solution, however, faces one crucial challenge on how to identify reasonable snippets under our weakly-supervised setting. The selection of hard snippets is non-trivial as there is no specific attention distribution pattern for them. For example, in Figure~\ref{fig:intro} baseline, \#3 hard snippet has a high response value while \#2 remains low. Noticing that ambiguous hard snippets are commonly found around boundary areas of the action instances, we propose a boundary-aware Hard Snippet Mining algorithm -- a simple yet effective importance sampling technique. Specifically, we first threshold T-CAS and then employ \textit{dilation} and \textit{erosion} operations temporally to mine the potential hard snippets. Since the hard snippets may either be action or background, we opt to distinguish them by their relative position. For easy snippets, they locate in the most discriminative parts, so snippets with \textit{top-k}/\textit{bottom-k} T-CAS scores are selected as easy action/background respectively. Moreover, we form two hard-easy contrastive pairs and conduct the feature refinement via the proposed SniCo Loss.

In a nutshell, the main contributions of this work are as follows: 
(1) Pioneeringly, we introduce the contrastive representation learning paradigm to WS-TAL and propose a SniCo Loss which effectively refines the feature representation of hard snippets. (2) A Hard Snippet Mining algorithm is proposed to locate potential hard snippets around boundaries, which serves as an efficient sampling strategy under our weakly-supervised setting. (3) Extensive experiments on THUMOS'14 and ActivityNet v1.2 datasets demonstrate the effectiveness of our proposed CoLA.
\begin{figure*}[!htbp]
\begin{center}
\includegraphics[width=0.9\linewidth]{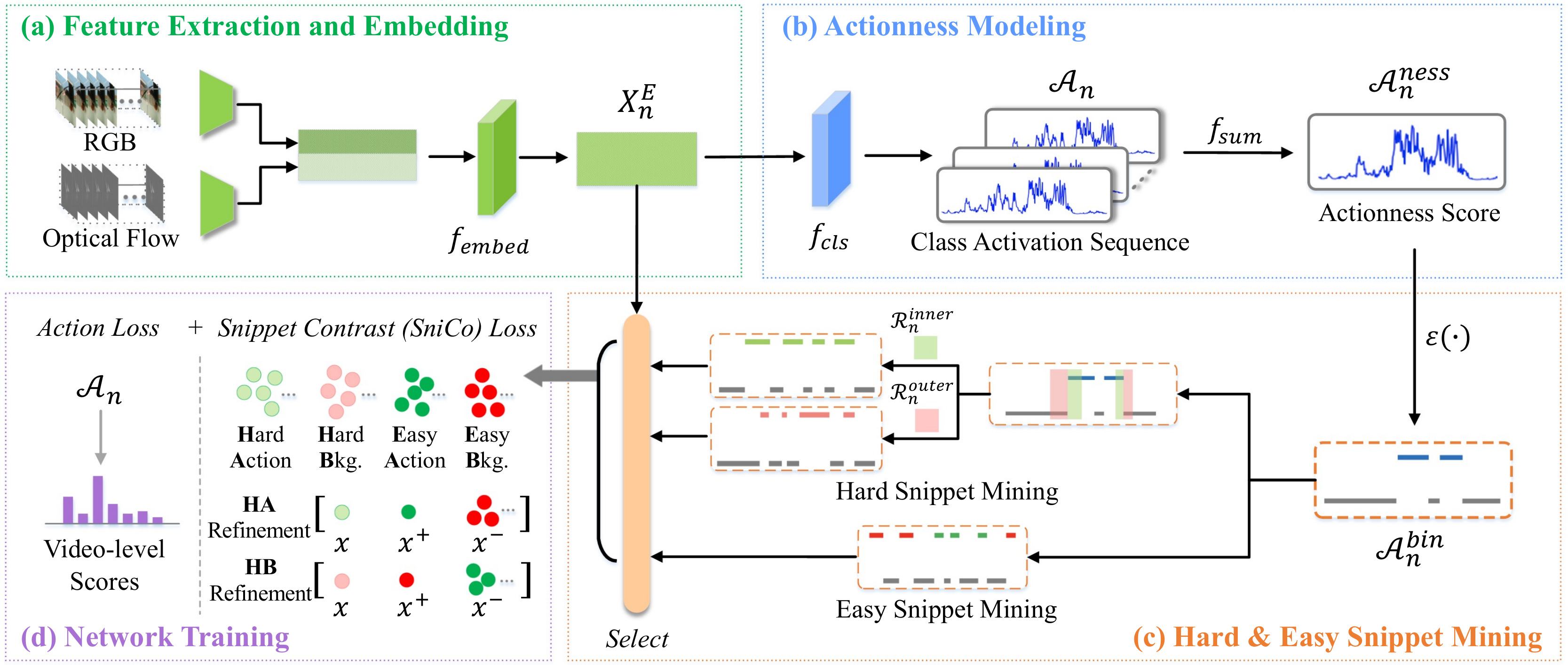}
\end{center}
   \caption{Illustration of the proposed CoLA, which consists of four parts: (a) Feature Extraction and Embedding to obtain the embedded feature $X_n^E$; (b) Actionness Modeling to gather class-agnostic action likelihood $\mathcal{A}^{ness}_n$; (c) Hard \& Easy Snippet Mining to select hard and easy snippets. (d) Network Training driven by Action Loss and Snippet Contrast (SniCo) Loss.}
\label{fig:short}
\end{figure*}

\section{Related Work}

\textbf{Fully-supervised Action Localization} utilizes frame-level annotations to locate and classify the temporal intervals of action instances from long untrimmed videos. Most existing works may be classified into two categories: proposal-based (top-down) and frame-based methods (bottom-up). Proposal-based methods~\cite{shou2016temporal,zhao2017temporal,xu2017r,dai2017temporal,chao2018rethinking,shou2017cdc,lin2018bsn,lin2019bmn,zeng2019graph,lin2020fast} first generate action proposals and then classify them as well as conduct temporal boundary regression. On the contrary, frame-based methods~\cite{lin2017single, buch2019end,long2019gaussian,zhao2020bottom} directly predict frame-level action category and location followed by some post-processing techniques.

\textbf{Weakly-Supervised Action Localization} only requires video-level annotations and has drawn extensive attention. UntrimmedNets~\cite{wang2017untrimmednets} address this problem by conducting the clip proposal classification first and then select relevant segments in a soft or hard manner. STPN~\cite{nguyen2018weakly} imposes a sparsity constraint to enforce the sparsity of the selected segments. Hide-and-seek~\cite{singh2017hide} and MAAN~\cite{yuan2019marginalized} try to extend the discriminative regions via randomly hiding patches or suppressing the dominant response, respectively. Zhong \textit{et al.}~\cite{zhong2018step} introduce a progressive generation procedure to achieve similar ends. W-TALC~\cite{paul2018w} applies the deep metric learning to be complementary with the Multiple Instance Learning formulation. 

\textbf{Discussion.} The single snippet cheating problem has not been fully studied though it is common in WS-TAL. Liu \textit{et al.}~\cite{liu2019completeness} pinpoint the action completeness modeling problem and the action-context separation problem. They develop a parallel multi-branch classification architecture with the help of the generated hard negative data. In contrast, our CoLA unifies these two problems and settles them in a lighter way with the proposed SniCo Loss. DGAM~\cite{shi2020weakly} mentions the \textit{action-context confusion issue}, \textit{i.e.,} context snippets near action snippets tend to be misclassified, which can be considered as a sub-problem of our \textit{single snippet cheating issue}. Besides, several background modeling works~\cite{nguyen2019weakly,lee2020background,shi2020weakly} can also be seen as one solution to this problem. Nguyen \textit{et al.}~\cite{nguyen2019weakly} utilizes an attention mechanism to model both foreground and background frame appearances and guide the generation of the class activation map. BaS-Net~\cite{lee2020background} introduces an auxiliary class for background and applies an asymmetrical training strategy to suppress the background snippet activation. However, these methods have inherent drawbacks as background snippets are not necessarily motionless and it is difficult to include them into one specific class. By contrast, our CoLA is a more adaptive and explainable solution to tackle these issues.

\textbf{Contrastive Representation Learning} uses data internal patterns to learn an embedding space where associated signals are brought together while unassociated ones are distinguished via Noise Contrastive Estimation (NCE)~\cite{gutmann2010noise}. CMC~\cite{tian2019contrastive} presents a contrastive learning framework that maximize mutual information between different views of the same scene to achieve a view-invariant representation. SimCLR~\cite{chen2020simple} selects the negative samples by using augmented views of other items in a minibatch. MoCo~\cite{he2020momentum} uses a momentum updated memory bank of old negative representations to get rid of the batch size restriction and enable the consistent use of negative samples. To our best knowledge, we are the first to introduce the noise contrastive estimation to WS-TAL task. Experiment results show that CoLA refines the hard snippet representation, thus benefiting the action localization.

\section{Method}
Generally, CoLA (shown in Figure~\ref{fig:short}) follows the feature extraction (Section~\ref{subsec:featureExtract}), actionness modeling (Section~\ref{subsec:acModel}) and hard \& easy snippet mining (Section~\ref{subsec:sampleSelect}) pipeline. The optimization loss terms and the inference process are detailed in Section~\ref{subsec:loss} and Section~\ref{subsec:infer}, respectively. 

\subsection{Feature Extraction and Embedding}
\label{subsec:featureExtract}
Assume that we are given a set of $N$ untrimmed videos $\{V_n\}_{n=1}^{N}$ and their video-level labels $\{y_n\}_{n=1}^{N}$, where $y_n \in \mathbb{R}^C$ is a multi-hot vector, and $C$ is the number of action categories. Following the common practice~\cite{nguyen2018weakly,nguyen2019weakly,lee2020background}, for each input untrimmed video $V_n$, we divide it into multi-frame non-overlapping $L_n$ snippets, \emph{i.e.}, $V_n = \{ S_{n,l} \}_{l=1}^{L_n}$. A fixed number of $T$ snippets $\{ S_{n,t} \}_{t=1}^{T}$ are sampled due to the variation of video length. Then the RGB features $X^{R}_n = \{ x_t^{R} \}_{t=1}^{T}$ and optical flow features $X^{O}_n = \{ x_t^{O} \}_{t=1}^{T}$ are extracted with pre-trained feature extractor (\emph{e.g.}, I3D \cite{carreira2017quo}), respectively. Here, $x_t^{R} \in \mathbb{R}^{d}$ and $x_t^{O} \in \mathbb{R}^{d}$, $d$ is the feature dimension of each snippet. Afterwards, we apply an embedding function $f_{embed}$ over the concatenation of $X^{R}_n$ and $X^{O}_n$ to obtain our extracted features $X_n^E \in \mathbb{R}^{T \times 2d}$. $f_{embed}$ is implemented with a temporal convolution followed by the ReLU activation function.

\subsection{Actionness Modeling} \label{subsec:acModel}
We introduce the concept \textit{Actionness} referring to the likelihood of containing a general action instance for each snippet. Before we specify the Actionness Modeling process, let’s revisit the commonly adopted Temporal Class Activation Sequence (T-CAS).

Given the embedded features $X_n^E$, we apply a classifier $f_{cls}$ to obtain snippet-level T-CAS. Specifically, the classifier contains a temporal convolution followed by ReLU activation and Dropout. This can be formulated as follows for a video $V_n$:
\begin{equation}
\mathcal{A}_n = f_{cls}(X_n^E; \phi_{cls}),
\label{equation:action_cls}
\end{equation}
\noindent where $\phi_{cls}$ represents the learnable parameters. The obtained $\mathcal{A}_n \in \mathbb{R}^{T \times C}$ represents the action classification results occurring at each temporal snippets.

Then, when it comes to modeling the actionness, one common way is to conduct the binary classification on each snippet, which yet will inevitably bring in extra overheads. Since the generated T-CAS $\mathcal{A}_n \in \mathbb{R}^{T \times C}$ in Eqn.~\ref{equation:action_cls} already contains snippet-level class-specific predictions, we simply sum T-CAS along the channel dimension ($f_{sum}$) followed by the Sigmoid function to obtain a class-agnostic aggregation and use it to represent the actionness $\mathcal{A}^{ness}_n \in \mathbb{R}^{T}$:
\begin{equation}
\mathcal{A}^{ness}_n = Sigmoid(f_{sum}(\mathcal{A}_n)).
\label{equation:aness}
\end{equation}

\subsection{Hard \& Easy Snippet Mining}
Recall that our aim is to use the easily spotted snippets as a priori to disambiguate controversial snippets. We systematically study the contrastive pair construction process for both hard and easy snippets.
\label{subsec:sampleSelect}
\subsubsection{Hard Snippet Mining}
\label{subsubsec:hardsampleSelect}
Intuitively, for most snippets located inside the action or background intervals, they are far from the temporal borders with less noise interference and have the relatively trustworthy feature representation. For boundary-adjacent snippets, however, they are less reliable because they are in the transitional areas between action and background, thus leading to ambiguous detection.

Base on the above observations, we argue that boundary-adjacent snippets can serve as the potential hard snippets under the weak supervision setting. Therefore, we build a novel Hard Snippet Mining algorithm to exploit hard snippets from the border areas. Then these mined hard snippets are divided into \textbf{hard action} and \textbf{hard background} according to their locations.

\begin{figure}[t]
\begin{center}
\includegraphics[width=1.0\linewidth]{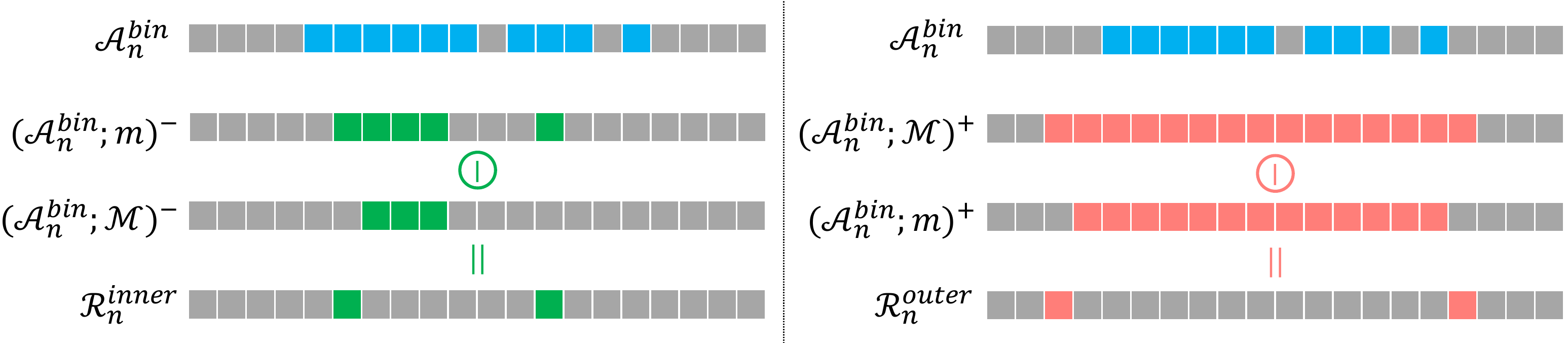}
\end{center}
   \caption{Illustration of the Hard Snippet Mining algorithm. \textbf{Left:} Subtract the eroded sequences with different masks to get the inner regions (green color); \textbf{Right:} Subtract the dilated sequences with different masks to get the outer regions (pink color). 
   }
\label{fig:pseudo_context_loc}
\end{figure}

Firstly, we threshold the actionness scores to generate a binary sequence (1 or 0 indicates the action or background location, respectively):
\begin{equation}
\mathcal{A}^{bin}_{n} = \varepsilon(\mathcal{A}^{ness}_n - \theta_b),
\label{equation:bin_mask}
\end{equation}
\noindent where $\varepsilon(\cdot)$ is the \emph{Heaviside step function} and $\theta_b$ is the threshold value, \emph{i.e.}, $\mathcal{A}_n^{bin}$ is 1 if $\mathcal{A}_n^{ness} \geq \theta_b$, 0 otherwise. Then, as shown in Figure~\ref{fig:pseudo_context_loc}, we apply two cascaded dilation or erosion operations to expand or narrow the temporal extent of action intervals. The differential areas with the diverse dilation or erosion degree are defined as the hard background or hard action regions: 
\begin{equation}
\begin{split}
\mathcal{R}_{n}^{inner} = (\mathcal{A}_{n}^{bin}; m)^- - (\mathcal{A}_{n}^{bin}; \mathcal{M})^-\\
\mathcal{R}_{n}^{outer} = (\mathcal{A}_{n}^{bin}; \mathcal{M})^+ - (\mathcal{A}_{n}^{bin}; m)^+ ,
\label{equation:outer_inner}
\end{split}
\end{equation}

\noindent where $(\cdot; *)^+$ and $(\cdot; *)^-$ represent the binary dilation and erosion operations with mask $*$, respectively. The inner region $\mathcal{R}_{n}^{inner}$ is defined as the different snippets between the eroded sequences with smaller mask $m$ and larger mask $\mathcal{M}$, as shown in Figure~\ref{fig:pseudo_context_loc} left part (in green color). Similarly, the outer region $\mathcal{R}_{n}^{outer}$ is calculated as the difference between the dilated sequences with larger mask $\mathcal{M}$ and smaller mask $m$, depicted in Figure~\ref{fig:pseudo_context_loc} right part (in pink color). 
Empirically, we regard the inner regions $\mathcal{R}_{n}^{inner}$ as hard action snippet sets since these regions are with $\mathcal{A}_{n}^{bin} = 1$. Similarly, the outer regions $\mathcal{R}_{n}^{outer}$ are considered as hard background snippet sets. Then the hard action snippets $X_n^{HA} \in \mathbb{R}^{k^{hard} \times 2d}$ are selected from $\mathcal{R}_{n}^{inner}$:
\begin{equation}
X_n^{HA} = \{ X^E_{n;t} | t \in \mathcal{I}_n^{act}, \mathcal{I}_n^{act} \in \mathcal{I}_{n}^{inner} \},
\label{equation:hard_action}
\end{equation}
\noindent where $\mathcal{I}_{n}^{inner}$ is the index set of snippets within $\mathcal{R}_{n}^{inner}$. $\mathcal{I}_n^{act}$ is the subset of $\mathcal{I}_{n}^{inner}$ with size $k^{hard}$
(\emph{i.e.}, $|\mathcal{I}_n^{act}| = k^{hard}$), and $k^{hard} = max(1, \lfloor \frac{T}{r^{hard}} \rfloor)$ is the hyper-parameter controlling the selected number of hard snippets, $r^{hard}$ is the sampling ratio. Considering the case that $k^{hard} > |\mathcal{I}_{n}^{inner}|$, we adopt \emph{sampling with replacement} mechanism to ensure the total $k^{hard}$ snippets can be selected. Similarly, the hard background snippets $X_n^{HB} \in \mathbb{R}^{k^{hard} \times 2d}$ are selected from $\mathcal{R}_{n}^{outer}$:
\begin{equation}
X_n^{HB} = \{ X^E_{n;t} | t \in \mathcal{I}_n^{bkg}, \mathcal{I}_n^{bkg} \in \mathcal{I}_{n}^{outer} \},
\label{equation:hard_background}
\end{equation}
\noindent where the notation definitions are similar to those in Eqn.~\ref{equation:hard_action} and we omit them for brevity. 

\subsubsection{Easy Snippet Mining} 
In order to form contrastive pairs, we still need to mine the discriminative easy snippets. Based on the well-trained fully-supervised I3D features, we hypothesize that the video snippets with top-\textit{k} and bottom-\textit{k} actionness scores are exactly easy action ($X_n^{EA} \in \mathbb{R}^{k^{easy} \times 2d}$) and easy background snippets ($X_n^{EB} \in \mathbb{R}^{k^{easy} \times 2d}$), respectively. Therefore, we conduct easy snippet mining based on the actionness scores calculated in Eqn.~\ref{equation:aness}. The specific process is as follows:
\begin{equation}
\begin{split}
X_n^{EA} = \{ X^E_{n;t} | t \in \mathcal{S}_n^{act}, t \not \in \mathcal{I}_n^{inner}, t \not \in \mathcal{I}_n^{outer}, \\ \mathcal{S}_n^{act} = \mathcal{S}_{n}^{DESC}[:k^{easy}] \} \\
X_n^{EB} = \{ X^E_{n;t} | t \in \mathcal{S}_n^{bkg}, t \not \in \mathcal{I}_n^{inner}, t \not \in \mathcal{I}_n^{outer}, \\ \mathcal{S}_n^{bkg} = \mathcal{S}_{n}^{ASC}[:k^{easy}] \},
\label{equation:easy_action_bkg}
\end{split}
\end{equation}
\noindent where $\mathcal{S}_{n}^{DESC}$ and $\mathcal{S}_{n}^{ASC}$ denotes the index of $\mathcal{A}_{n}^{ness}$ sorting by DESC and ASC order respectively. $k^{easy} = max(1, \lfloor \frac{T}{r^{easy}} \rfloor)$, $r^{easy}$ is a hyper-parameter representing the selection ratio. Note that we remove the snippets in the hard snippet areas $\mathcal{R}_n^{inner}$ and $\mathcal{R}_n^{outer}$ to avoid conflict.

\subsection{Network Training} \label{subsec:loss}
Based on the mined hard and easy snippets, our CoLA introduces an additional Snippet Contrast (SniCo) Loss ($\mathcal{L}_{s}$) and achieves considerable improvement compared with the \textit{baseline} model. The total loss can be represented as follows:
\begin{equation}
\mathcal{L}_{total} = \mathcal{L}_{a} + \lambda \mathcal{L}_{s},
\label{equation:final_loss}
\end{equation}
where $\mathcal{L}_{a}$ and $\mathcal{L}_{s}$ denote the Action Loss and the SniCo Loss, respectively. $\lambda$ is the balance factor. We elaborate on these two terms as follows.

\subsubsection{Action Loss} \label{method:act_loss}
Action Loss ($\mathcal{L}_{a}$) is the classification loss between the predicted video category and the ground truth. To get the video-level predictions, we aggregate snippet-level class scores computed in Eqn.~\ref{equation:action_cls}. Following~\cite{wang2017untrimmednets,paul2018w,lee2020background}, we take the \textit{top-k mean} strategy: for each class $c$, we take $k^{easy}$ terms with the largest class-specific T-CAS values and compute their means as $a_{n;c}$, namely the video-level class score for class $c$ of video $V_n$. After obtaining $a_{n;c}$ for all the $C$ classes, we apply a Softmax function on $a_n$ along the class dimension to get the video-level class possibilities $p_n \in \mathbb{R}^{C}$. Action Loss ($\mathcal{L}_{a}$) is then calculated in the cross-entropy form: 
\begin{equation}
\mathcal{L}_{a} = -\frac{1}{N} \sum^{N}_{n=1} \sum^{C}_{c=1} \hat{y}_{n;c}  \mathrm{log} (p_{n;c}),
\label{equation:action_loss}
\end{equation}
where $\hat{y}_n \in \mathbb{R}^{C}$ is the normalized ground-truth.

\subsubsection{Snippet Contrast (SniCo) Loss}
Contrastive learning has been used on image or patch levels~\cite{bachman2019learning,henaff2019data}. For our application, given the extracted feature embedding $X_n^E$, the contrastive learning is applied in the snippet level. We name it \textit{Snippet Contrast (SniCo)} Loss ($\mathcal{L}_{s}$), which  aims to refine the snippet-level feature of hard snippets and obtain a more informative feature distribution. Considering that the hard snippets are classified as hard action and hard background, we form two contrastive pairs in $\mathcal{L}_{s}$ accordingly, namely ``HA refinement" and ``HB refinement'', where HA and HB 
are short for hard action and hard background respectively. ``HA refinement" aims to transform the hard action snippet features by driving hard action and easy action snippets compactly in feature space and ``HB refinement" is similar. 

Formally, the query $\boldsymbol{x} \in \mathbb{R}^{1 \times 2 d}$, positive $\boldsymbol{x}^{+} \in \mathbb{R}^{1 \times 2 d}$, and $S$ negatives $\boldsymbol{x}^{-} \in \mathbb{R}^{S \times 2d}$ are selected from pre-mined snippets. As shown in Figure~\ref{fig:short}(d), for ``HA refinement", $\boldsymbol{x} \sim X_n^{HA}, \boldsymbol{x}^{+} \sim X_n^{EA}$,  $\boldsymbol{x}^{-} \sim X_n^{EB}$; for ``HB refinement", $\boldsymbol{x} \sim X_n^{HB}, \boldsymbol{x}^{+} \sim X_n^{EB}$,  $\boldsymbol{x}^{-} \sim X_n^{EA}$. We project them to a normalized unit sphere to prevent the space from collapsing or expanding. An $(S+1)$ -way classification problem using the cross-entropy loss is set up to represent the probability of the positive example being selected over negatives. Following~\cite{he2020momentum}, we compute the distances between the query and other examples with a temperature scale $\tau = 0.07$:

{\setlength\abovedisplayskip{1pt}
\setlength\belowdisplayskip{1pt}
\begin{equation}
\begin{aligned}
\ell( & \boldsymbol{x}, \boldsymbol{x^{+}}, \boldsymbol{x^{-}} )  \\=
&-\log \left[\frac{\exp \left(\boldsymbol{x}^{\mathrm{T}} \cdot \boldsymbol{x}^{+} / \tau \right)}{\exp \left(\boldsymbol{x}^{\mathrm{T}} \cdot \boldsymbol{x}^{+} / \tau \right)+\sum_{s=1}^{S} \exp \left(\boldsymbol{x}^{\mathrm{T}} \cdot \boldsymbol{x}^{-}_s / \tau \right)}\right],
\end{aligned}
\end{equation}}

\noindent where $\boldsymbol{x}^{\mathrm{T}}$ is the transpose of $\boldsymbol{x}$ and the proposed SniCo Loss is as follows:

\begin{equation}
\begin{aligned}
\mathcal{L}_{s} &=\underbrace{\mathbb{E}_{\boldsymbol{x} \sim X_{n}^{H A}, \boldsymbol{x^{+}} \sim X_{n}^{E A}, \boldsymbol{x^{-}} \sim X_{n}^{E B}} \ell\left(\boldsymbol{x}, \boldsymbol{x^{+}}, \boldsymbol{x^{-}}\right) }_\text{HA refinement}\\
&+\underbrace{\mathbb{E}_{\boldsymbol{x} \sim X_{n}^{H B}, \boldsymbol{x^{+}} \sim X_{n}^{E B}, \boldsymbol{x^{-}} \sim X_{n}^{E A}} \ell\left(\boldsymbol{x}, \boldsymbol{x^{+}}, \boldsymbol{x^{-}}\right)}_\text{HB refinement},
\end{aligned}
\label{equation:cc_loss}
\end{equation}

\noindent where $S$ represents the number of negative snippets and $\boldsymbol{x}_{s}^{-} \in \mathbb{R}^{2d}$ means the \textit{s}-th negative. In this way, we maximize \textit{mutual information} between the easy and hard snippets of the same category (action or background), which helps refine the feature representation and thereby alleviating the single snippet cheating issue.

\subsection{Inference} \label{subsec:infer}

Given an input video, we first predict its snippet-level class activations to form T-CAS and aggregate top-$k^{easy}$ scores described in Sec.~\ref{method:act_loss} to get the video-level predictions. Then the categories with scores larger than $\theta_{v}$ are selected for further localization. For each selected category, we threshold its corresponding T-CAS with $\theta_{s}$ to obtain candidate video snippets. Finally, continuous snippets are grouped into proposals and Non-Maximum Suppression (NMS) is applied to remove duplicated proposals.
\section{Experiments}

\subsection{Datasets}

We evaluate our CoLA on two popular action localization benchmark datasets including THUMOS'14~\cite{idrees2017thumos} and ActivityNet v1.2~\cite{caba2015activitynet}. We only use the video-level category labels for network training.

\textbf{THUMOS'14} includes untrimmed videos with 20 categories. The video length varies greatly and each video may contain multiple action instances. By convention~\cite{lee2020background,shi2020weakly}, we use the 200 videos in validation set for training and the 213 videos in testing set for evaluation.

\textbf{ActivityNet v1.2} is a popular large-scale benchmark for TAL with 100 categories. Following the common practice~\cite{wang2017untrimmednets,shou2018autoloc}, we train on the training set with 4,819 videos and test on the validation set with 2,383 videos.

\subsection{Implementation Details}

\textbf{Evaluation Metrics.} We follow the standard evaluation protocol by reporting mean Average Precision (mAP) values under different intersection over union (IoU) thresholds. The evaluation on both datasets are conducted using the benchmark code provided by ActivityNet\footnote{https://github.com/activitynet/ActivityNet/}.

\textbf{Feature Extractor.} We use I3D~\cite{carreira2017quo} network pre-trained on Kinetics~\cite{carreira2017quo} for feature extraction. Note that the I3D feature extractor is not fine-tuned for fair comparison. TVL1~\cite{perez2013tv} algorithm is applied to extract optical flow stream from RGB stream in advance. Each video stream is divided into 16-frame non-overlapping snippets and the snippet-wise RGB and optical flow features are with 1024-dimension.

\textbf{Training Details. } The number of sampled snippets $T$ for THUMOS’14 and ActivityNet v1.2 is set to 750 and 50, respectively. All hyper-parameters are determined by grid search: $r^{easy}=5$, $r^{hard}=20$, $S=k^{easy} = max(1, \lfloor \frac{T}{r^{easy}} \rfloor)$. We set $\lambda=0.01$ in Eqn.~\ref{equation:final_loss}. $\theta_b$ in Eqn.~\ref{equation:bin_mask} is set to 0.5 for both datasets. Dilation and erosion masks $\mathcal{M}$ and $m$ are set to 6 and 3 in our experiments. We utilize Adam optimizer with a learning rate of $1e-4$. We train for total 6k epochs with a batch size of 16 for THUMOS'14 and for total 8k epochs with a batch size of 128 for ActivityNet v1.2.

\textbf{Testing Details. } We set $\theta_v$ to 0.2 and 0.1 for THUMOS’14 and ActivityNet v1.2, respectively. For proposal generation, we use multiple thresholds that $\theta_s$ is set as [0:0.25:0.025] for THUMOS'14 and [0:0.15:0.015] for ActivityNet v1.2, then Non-Maximum Suppression (NMS) is performed with IoU threshold 0.6.

\begin{table*}[t]
\begin{center}
\caption{Comparisons with state-of-the-art TAL methods on THUMOS'14 dataset. The mAP values at different IoU thresholds are reported. The AVG column shows the averaged mAP under the thresholds [0.1:0.7:0.1]. UNT is the abbreviation for UntrimmedNet feature.}
\label{table:thumos_comp}
\resizebox{0.83\linewidth}{!}{
\begin{threeparttable}
\begin{tabular}{ccccccccccc}
\toprule
\multirow{2.5}*{\textbf{\tabincell{c}{Supervision\\(Feature)}}} & \multirow{2.5}*{\textbf{Method}} & \multirow{2.5}*{\textbf{Publication}} & \multicolumn{8}{c}{\textbf{mAP@IoU (\%)}}\\
\cmidrule(lr){4-11}
& & & 0.1 & 0.2 & 0.3 & 0.4 & 0.5 & 0.6 & 0.7 & AVG\\
\midrule
\multirow{5}*{\tabincell{c}{Full\\(-)}} & R-C3D~\cite{xu2017r} & \emph{ICCV 2017} & 54.5 & 51.5 & 44.8 & 35.6 & 28.9 & - & - & - \\
& SSN~\cite{zhao2017temporal}  & \emph{ICCV 2017} & 66.0 & 59.4 & 51.9 & 41.0 & 29.8 & - & - & - \\
& TAL-Net~\cite{chao2018rethinking} & \emph{CVPR 2018} & 59.8 & 57.1 & 53.2 & 48.5 & 42.8 & 33.8 & 20.8 & 45.1\\
& P-GCN~\cite{zeng2019graph} & \emph{ICCV 2019} & 69.5 & 67.8 & 63.6 & 57.8 & 49.1 & - & - & - \\
& G-TAD~\cite{xu2020g} & \emph{CVPR 2020} & - & - & \textbf{66.4} & \textbf{60.4} & \textbf{51.6} & \textbf{37.6} & \textbf{22.9} & - \\
\midrule
\midrule
\multirow{3}*{\tabincell{c}{Weak\\(-)}} & Hide-and-Seek~\cite{singh2017hide} & \emph{ICCV 2017} & 36.4 & 27.8 & 19.5 & 12.7 & 6.8 & - & - & - \\
& UntrimmedNet~\cite{wang2017untrimmednets} & \emph{CVPR 2017} & 44.4 & 37.7 & 28.2 & 21.1 & 13.7 & - & - & -\\
& Zhong \emph{et al.}~\cite{zhong2018step} & \emph{ACMMM 2018} & 45.8 & 39.0 & 31.1 & 22.5 & 15.9 & - & - & -\\
\hdashline
\multirow{3}*{\tabincell{c}{Weak\\(UNT)}} & AutoLoc~\cite{shou2018autoloc} & \emph{ECCV 2018} & - & - & 35.8 & 29.0 & 21.2 & 13.4 & 5.8 & - \\
& CleanNet~\cite{liu2019weakly} & \emph{ICCV 2019} & - & - & 37.0 & 30.9 & 23.9 & 13.9 & 7.1 & - \\
& Bas-Net~\cite{lee2020background} & \emph{AAAI 2020} & - & - & 42.8 & 34.7 & 25.1 & 17.1 & 9.3 & - \\
\hdashline
\multirow{9}*{\tabincell{c}{Weak\\(I3D)}} & STPN~\cite{nguyen2018weakly} & \emph{CVPR 2018} & 52.0 & 44.7 & 35.5 & 25.8 & 16.9 & 9.9 & 4.3 & 27.0\\
& Liu \emph{et al.}~\cite{liu2019completeness}  & \emph{CVPR 2019} & 57.4 & 50.8 & 41.2 & 32.1 & 23.1 & 15.0 & 7.0 & 32.4\\
& Nguyen \emph{et al.}~\cite{nguyen2019weakly} & \emph{ICCV 2019} & 60.4 & 56.0 & 46.6 & 37.5 & 26.8 & 17.6 & 9.0 & 36.3\\
& BaS-Net~\cite{lee2020background} & \emph{AAAI 2020} & 58.2 & 52.3 & 44.6 & 36.0 & 27.0 & 18.6 & 10.4 & 35.3\\
& DGAM~\cite{shi2020weakly} & \emph{CVPR 2020} & 60.0 & 54.2 & 46.8 & 38.2 & 28.8 & 19.8 & 11.4 & 37.0\\
& ActionBytes~\cite{Jain_2020_CVPR} & \emph{CVPR 2020} & - & - & 43.0 & 35.8 & 29.0 & - & 9.5 & -\\
& A2CL-PT~\cite{min2020adversarial} & \emph{ECCV 2020} & 61.2 & 56.1 & 48.1 & 39.0 & 30.1 & 19.2 & 10.6 & 37.8\\
& TSCN~\cite{zhai2020two} & \emph{ECCV 2020} & 63.4 & 57.6 & 47.8 & 37.7 & 28.7 & 19.4 & 10.2 & 37.8\\
& \textbf{CoLA (Ours)} \textcolor{red}{\textsuperscript{$\dagger$}} & - & \textbf{66.2} & \textbf{59.5} & \textbf{51.5} & \textbf{41.9} & \textbf{32.2} & \textbf{22.0} & \textbf{13.1} &  \textbf{40.9}\\
\bottomrule
\end{tabular}
\begin{tablenotes}
\small
\item \textcolor{red}{$\dagger$} After fine-tuning some hyper-parameter settings, the experimental results are better than these, please see \href{https://github.com/zhang-can/CoLA}{our code} for details.
\end{tablenotes}
\end{threeparttable}
}
\end{center}
\end{table*}

\begin{table}[t]
\begin{center}
\vspace{-6pt}
\caption{Comparison results on ActivityNet v1.2 dataset. The AVG column shows the averaged mAP under the thresholds [0.5:0.95:0.05]. UNT and I3D are abbreviations for UntrimmedNet feature and I3D feature, respectively.}
\label{table:anet12_comp}
\resizebox{0.9\linewidth}{!}{
\begin{tabular}{ccccccc}
\toprule
\multirow{2.5}*{\textbf{Sup.}} & \multirow{2.5}*{\textbf{Method}} & \multicolumn{4}{c}{\textbf{mAP@IoU (\%)}}\\
\cmidrule(lr){3-7}
& & 0.5 & 0.75 & 0.95 & AVG \\
\midrule
Full & SSN~\cite{zhao2017temporal} & 41.3 & 27.0 & 6.1 & 26.6\\
\midrule
Weak & UntrimmedNet~\cite{wang2017untrimmednets} & 7.4 & 3.2 & 0.7 & 3.6\\
(UNT)& AutoLoc~\cite{shou2018autoloc} & 27.3 & 15.1 & 3.3 & 16.0\\
\hdashline
\multirow{8}*{\tabincell{c}{Weak\\(I3D)}} & W-TALC~\cite{paul2018w} & 37.0 & 12.7 & 1.5 & 18.0\\
& TSM~\cite{yu2019temporal} & 28.3 & 17.0 & 3.5 & 17.1\\
& CleanNet~\cite{liu2019weakly} & 37.1 & 20.3 & 5.0 & 21.6\\
& Liu \emph{et al.}~\cite{liu2019completeness} & 36.8 & 22.0 & 5.6 & 22.4\\
& BaS-Net~\cite{lee2020background} & 38.5 & 24.2 & 5.6 & 24.3\\
& DGAM~\cite{shi2020weakly} & 41.0 & 23.5 & 5.3 & 24.4\\
& TSCN~\cite{zhai2020two} & 37.6 & 23.7 & 5.7 & 23.6\\
& \textbf{CoLA (Ours)} & \textbf{42.7} & \textbf{25.7} & \textbf{5.8} & \textbf{26.1}\\
\bottomrule
\end{tabular}
}
\end{center}
\end{table}

\subsection{Comparison with State-of-the-Arts}

We compare our CoLA with the state-of-the-art fully-supervised and weakly-supervised TAL approaches on THUMOS'14 testing set. As shown in Table~\ref{table:thumos_comp}, CoLA achieves the impressive performance, \textit{i.e.,} we consistently outperform previous weakly-supervised methods at all IoU thresholds. Specifically, our method achieves 32.2\% mAP@0.5 and 40.9\% mAP@AVG, bringing the state-of-the-art to a new level. Notably, even with a much lower level of supervision, our method is even comparable with several fully-supervised methods, following the latest fully-supervised approaches with the least gap.

We also conduct experiments on ActivityNet v1.2 validation set and the comparison results are summarized in Table~\ref{table:anet12_comp}. Again, our method shows significant improvements over state-of-the-art weakly-supervised TAL methods while maintaining competitive compared with other fully-supervised methods. The consistent superior results on both datasets signify the effectiveness of CoLA.

\subsection{Ablation Studies}
\label{subsec:abExp}
\begin{table}[!t]
\begin{center}
\caption{Ablation analysis on loss terms on THUMOS'14.}
\label{table:abla_contrast}
\resizebox{0.9\linewidth}{!}{
\begin{tabular}{ccc}
\toprule
\textbf{Setting} & \textbf{Loss} & \textbf{mAP@0.5} ($\bm \Delta$)\\
\midrule
CoLA (Ours) & $\mathcal{L}_{a}+\mathcal{L}_{s}$ & \textbf{32.2\%} \\
baseline & $\mathcal{L}_{a}$ & 24.7\% (-7.5\%) \\
\hdashline
CoLA w/o HB ref. & $\mathcal{L}_{a}+\mathcal{L}_{s}^{HA}$ & 29.7\% (-2.5\%) \\
CoLA w/o HA ref. & $\mathcal{L}_{a}+\mathcal{L}_{s}^{HB}$ & 30.4\% (-1.8\%) \\
\bottomrule
\end{tabular}
}
\end{center}
\vspace{-10pt}
\end{table}

\begin{figure}[!t]
\begin{center}
\includegraphics[width=0.9\linewidth]{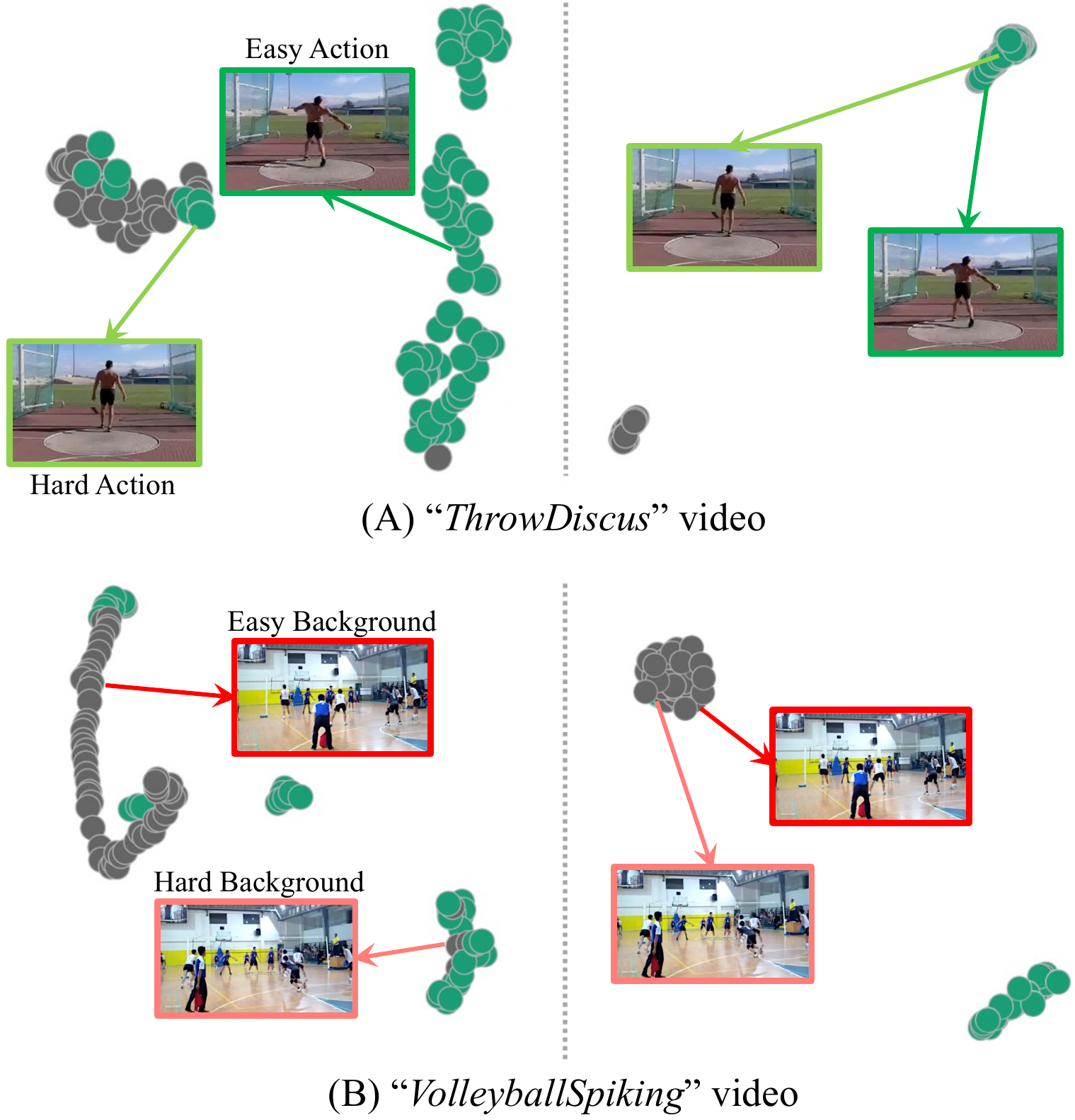}
\end{center}
\vspace{-6pt}
   \caption{UMAP visualizations of feature embeddings $X_{n}^{E}$. Left: baseline; Right: CoLA. Green points represent action embeddings and gray points denote background embeddings. CoLA achieves a more separable feature distribution compared to baseline.}
\label{fig:feats_vis}
\end{figure}

In this section, we conduct multiple ablation studies to provide more insights about our design intuition. By convention~\cite{nguyen2019weakly,shi2020weakly,lee2020background}, all the ablation experiments are performed on the THUMOS'14 testing set.

\textbf{Q1: How does the proposed SniCo Loss help?} To evaluate the effectiveness of our SniCo Loss ($\mathcal{L}_{s}$), we conduct a comparison experiment with only the action loss $\mathcal{L}_a$ as supervision, namely \textit{baseline} in Table~\ref{table:abla_contrast}. The statistical results in Table~\ref{table:abla_contrast} demonstrate that by introducing $\mathcal{L}_{s}$, the performance largely gains by 7.5\% in mAP@0.5, partially because SniCo Loss effectively guides the network to achieve better feature distribution tailored for WS-TAL. To illustrate this, we randomly select 2 videos from THUMOS'14 testing set and calculate the feature embeddings $X_{n}^{E}$ for baseline and CoLA, respectively. These embeddings are then projected to 2-dimensional space using UMAP~\cite{2018arXivUMAP}, as shown in Figure~\ref{fig:feats_vis}. Notice that compared with baseline, SniCo Loss helps to separate the action and background snippets more precisely, especially for those ambiguous hard snippets. Overall, the above analyses strongly justify the significance of our proposed SniCo Loss. 

\begin{figure}[t]
\begin{center}
\includegraphics[width=0.8\linewidth]{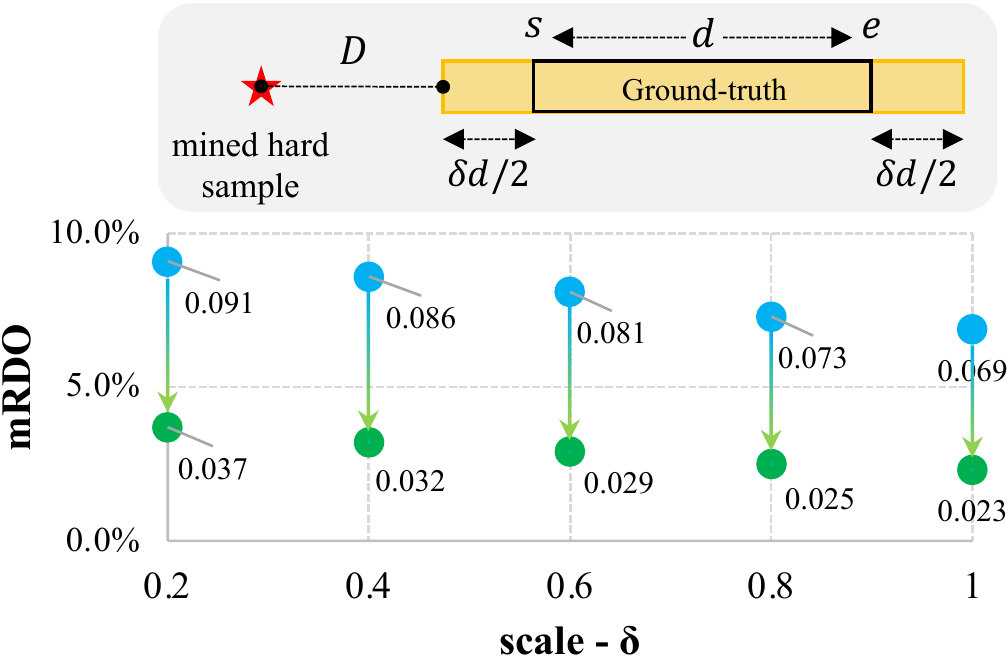}
\end{center}
   \caption{Effectiveness verification of Hard Snippet Mining algorithm. \textbf{Top:} Illustration of relative distance offsets (RDO) for a mined snippet. \textbf{Bottom:} The mean RDO (mRDO) \textit{vs.} different scales $\delta$ at epoch 0 (blue) and epoch 2k (green). }
   \vspace{-6pt}
\label{fig:abla_context_loc}
\end{figure}

\textbf{Q2: Is it necessary to consider both HA and HB refinements in SniCo Loss?}
To explore this, we conduct ablated experiments with two variants of SniCo Loss, each of which contains only one type of refinement in Eqn.~\ref{equation:cc_loss}, namely $\mathcal{L}_{s}^{HA}$ and $\mathcal{L}_{s}^{HB}$, respectively. Table~\ref{table:abla_contrast} shows that the performance drops dramatically with either kind of refinement removed, suggesting that both refinements contribute to the improved performance. 

\textbf{Q3: Are our mined hard snippets meaningful?} How to evaluate the effectiveness of the mined hard snippets is nontrivial. As discussed in Sec.~\ref{subsubsec:hardsampleSelect}, indistinguishable frames usually exist within or near the action temporal intervals, so we define such temporal areas as \textit{error-prone regions}. Specifically, given a ground-truth action instance with interval $[s,e]$ and duration $d=e-s$, we define its $\delta$-scale error-prone regions as $[s-\delta \frac{d}{2},e+\delta \frac{d}{2}]$, as illustrated in Figure~\ref{fig:abla_context_loc} top part. Then, to evaluate the positional relationship of our mined hard snippets with the error-prone areas, \emph{relative distance offset} (RDO) is defined as follows: 1) if a mined hard snippet does not fall into any of the error-prone regions, $RDO=\frac{D}{T}$, where $D$ is the nearest distance between this snippet and all error-prone regions, and $T$ is the video length; 2) otherwise, $RDO=0$. As shown in Figure~\ref{fig:abla_context_loc} bottom part, the mean RDO values (mRDO) of all the videos are evaluated under different scales $\delta$ at two training snapshots(epoch 0 and epoch 2k). The mRDO consistently drops at all scales $\delta$, indicating that our mined hard snippets are captured more precisely as the training goes on. Even under the most stringent condition ($\delta=0.2$), the mRDO is only 3.7\%, which suggests that most of our mined hard snippets locate in such error-prone areas and thus contribute to the network training.

\begin{table}[!t]
\begin{center}
\caption{Ablation analysis on the negative sample size $S$.}
\label{table:abla_S}
\resizebox{0.9\linewidth}{!}{
\begin{tabular}{cccccccc}
\toprule
\textbf{$S$} & 1 & 4 & 16 & 64 & 125 ($k^{easy}$)\\
\midrule
\textbf{mAP@0.5} & 28.9 & 30.4 & 31.3 & 31.9 & \textbf{32.2}\\
\bottomrule
\end{tabular}
}
\end{center}
\vspace{-6pt}
\end{table}

\textbf{Evaluation on the negative sample size $S$.} 
Table~\ref{table:abla_S} reports the experimental results evaluated with different negative sample sizes $S$. According to Eqn.~\ref{equation:cc_loss}, negative snippets are randomly chosen from the mined easy snippets, so $S \leq k^{easy}$. As shown, the mAP value is positively correlated with $S$, indicating that contrastive power increases by adding more negatives. This phenomenon is consistent with many self-supervised contrastive learning works~\cite{oord2018representation,he2020momentum,chen2020simple} and a recent supervised one~\cite{khosla2020supervised}, which partially verifies the efficacy of our hard and easy snippet mining algorithm for weakly-supervised TAL task.

\begin{table}[!t]
\begin{center}
\caption{Ablation analysis on the mask size $\mathcal{\bm M}$ and $m$.}
\label{table:abla_mask}
\resizebox{0.9\linewidth}{!}{
\begin{tabular}{ccccccc}
\toprule
\textbf{$\mathcal{\bm M}$}($m=3$) & 4 & 5 & 6 & 7 & 8 & 9\\
\midrule
\textbf{mAP@0.5} & 30.9 & 31.8 & \textbf{32.2} & 32.0 & 31.8 & 32.1 \\
\midrule
\midrule
\textbf{$\bm m$}($\mathcal{\bm M}=6$) & 0 & 1 & 2 & 3 & 4 & 5\\
\midrule
\textbf{mAP@0.5} & 30.3 & 31.7 & 32.0 & \textbf{32.2} & 32.0 & 31.9 \\
\bottomrule
\end{tabular}
}
\end{center}
\vspace{-10pt}
\end{table}

\textbf{Evaluation on the mask size $\mathcal{M}$ and $\bm m$.} 
We have defined two operation degrees (with larger $\mathcal{M}$ and smaller $m$) for temporal interval erosion and dilation in Eqn.~\ref{equation:outer_inner}. Here we seek to evaluate the effect of different mask sizes. For simplification, we first fix $m=3$ and vary $\mathcal{M}$ from 4 to 9, then we fix $\mathcal{M}=6$ and change $m$ from 0 to 5. The results are shown in Table~\ref{table:abla_mask}. The best result is achieved when setting $\mathcal{M}=6$ and $m=3$. 
Besides, it is quite evident that the performance remains stable across a wide range of $\mathcal{M}$ and $m$, demonstrating the robustness of our proposed Hard Snippet Mining algorithm.

\subsection{Qualitative Results}

\begin{figure}[t]
\begin{center}
\includegraphics[width=1.0\linewidth]{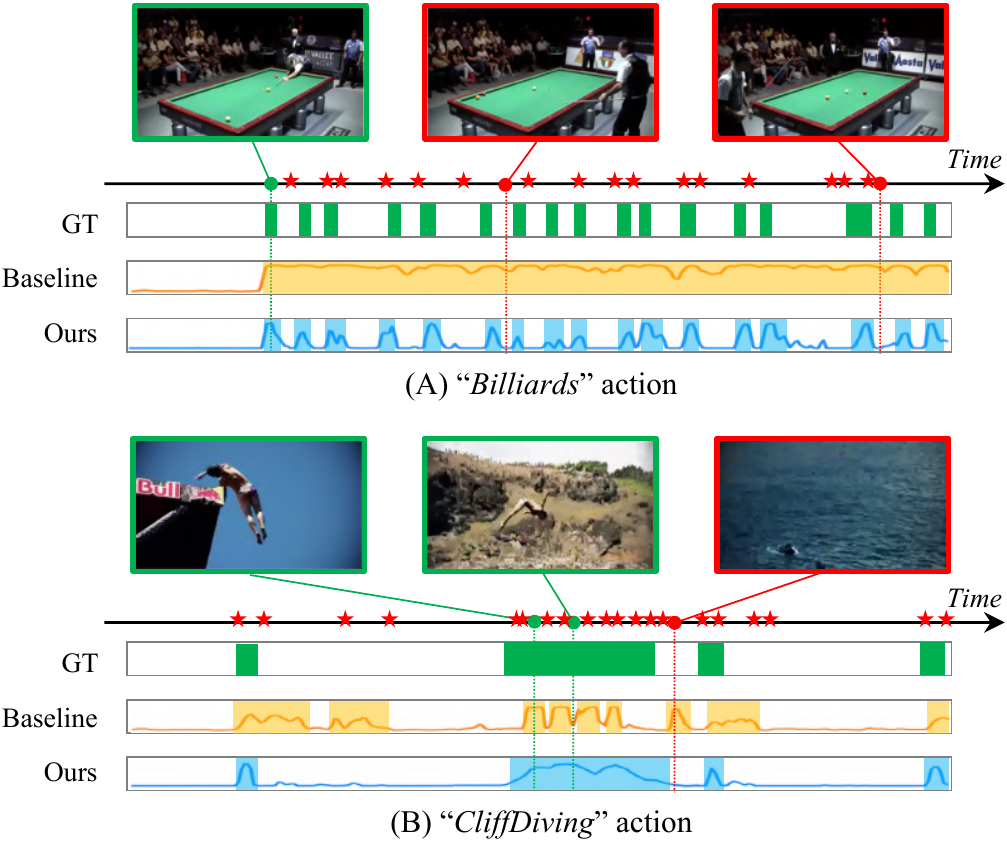}
\end{center}
\vspace{-6pt}
   \caption{Qualitative comparisons with baseline on THUMOS'14. For baseline and CoLA, we visualize the one-dimensional T-CAS and the localized regions. For clarity, frames with green bounding boxes refer to ground-truth actions and those in red refer to ground-truth backgrounds. Red pentagrams along the time axis denote the mined hard snippet locations (computed at epoch 2k).}
\vspace{-6pt}
\label{fig:vis_res}
\end{figure}

We visualize T-CAS results for two actions on THUMOS'14 in Figure~\ref{fig:vis_res}. Our CoLA has a more informative T-CAS distribution compared to baseline, thus leading to more accurate localization. Figure~\ref{fig:vis_res}-A depicts a typical case that all the frames in a video share the similar elements, \emph{i.e.}, humans, billiard table and balls. By introducing SniCo Loss, our method can seek the subtle differences between action and hard background, thereby avoiding many false positives produced by single Action Loss (baseline). Figure~\ref{fig:vis_res}-B demonstrates a ``CliffDiving" action observed from different camera views. The baseline method fails to localize the complete interval and outputs short and sparse prediction results. Our method successfully identifies the entire ``CliffDiving'' action and suppress the false positive detections. We also visualize the mined hard snippet locations (computed at epoch 2k) on the time axis (marked as red pentagram). As expected, these snippets are misclassified in baseline and CoLA refines their representation to achieve better performance. This visualization also helps explain \textbf{Q3} in Section~\ref{subsec:abExp}. For more visualization results, please refer to our supplementary materials.

\section{Conclusion}
In this paper, we have proposed a novel framework (CoLA) to address the single snippet cheating issue in weakly-supervised action localization. We leverage the intuition that hard snippets frequently lay in the boundary regions of the action instances and propose a Hard Snippet Mining algorithm to localize them. Then we apply a SniCo Loss to refine the feature representation of the mined hard snippets with the help of easy snippets which locate in the most discriminative regions. Experiments conducted on two benchmarks including THUMOS'14 and ActivityNet v1.2 have validated the state-of-the-art performance of CoLA. 
\vspace{10pt}

{\footnotesize \noindent \textbf{Acknowledgements.} This paper was partially supported by the IER foundation (No. HT-JD-CXY-201904) and Shenzhen Municipal Development and Reform Commission (Disciplinary Development Program for Data Science and Intelligent Computing). Special acknowledgements are given to Aoto-PKUSZ Joint Lab for its support.}

{\small
\bibliographystyle{ieee_fullname}

}

\end{document}